\documentclass{article}

\PassOptionsToPackage{numbers, compress}{natbib}
\usepackage[preprint]{neurips_2026}

\usepackage[utf8]{inputenc}
\usepackage[T1]{fontenc}
\usepackage{hyperref}
\usepackage{url}
\usepackage{booktabs}
\usepackage{amsfonts}
\usepackage{amsmath}
\usepackage{amssymb}
\usepackage{nicefrac}
\usepackage{microtype}
\usepackage{xcolor}
\usepackage{graphicx}
\usepackage{algorithm}
\usepackage{algpseudocode}
\usepackage{multirow}
\usepackage{enumitem}
\usepackage{bbding}        % \CheckmarkBold and \XSolidBrush
\usepackage{colortbl}      % \columncolor for highlighted columns
\usepackage{subcaption}
\usepackage{adjustbox}
\usepackage{multirow}
\usepackage{adjustbox}
\usepackage{makecell}
\usepackage{tabularx}
\usepackage[table]{xcolor}
\usepackage{array}
\usepackage[most]{tcolorbox}
\usepackage{courier} % for \fontfamily{pcr}
\usepackage{fancyvrb}

\newtcblisting{promptbox}[2][]{
    breakable,
    enhanced,
    colback=customwhite,
    colframe=customgreen,
    title=#2,
    listing only,
    listing engine=fancyvrb,
    listing options={
        fontsize=\scriptsize,
        breaklines=true,
        breakanywhere=true,
        breaksymbolleft={},
        breaksymbolright={},
        tabsize=2
    },
    #1
}

\usepackage{xspace}

\definecolor{customwhite}{HTML}{FFFFFF}
\definecolor{customgreen}{HTML}{2E8B57}
\definecolor{customblue}{HTML}{4169E1}
\definecolor{custompurple}{HTML}{7B68EE}
\title{Revisiting DAgger in the Era of LLM-Agents}

\author{Changhao Li$^1$, Rushi Qiang$^1$$^\dagger$, Jiawei Huang$^2$$^\dagger$, Chenxiao Gao$^1$$^\dagger$,\\ 
\textbf{Chao Zhang$^1$, Niao He$^{2}$, Bo Dai$^{1}$} \\
$^\dagger$Equal Second Authorship, $^1$Georgia Institute of Technology, $^2$ETH Zurich\\
\texttt{\{cli911, rqiang6, cgao\}@gatech.edu, jiawei.huang@inf.ethz.ch}\\
\texttt{chaozhang@gatech.edu, niao.he@inf.ethz.ch, bodai@cc.gatech.edu} \\
}

\begin{document}

\maketitle

% =========================================================================
\begin{abstract}
Long-horizon LM agents learn from multi-turn interaction, where a single early mistake can alter the subsequent state distribution and derail the whole trajectory. Existing recipes fall short in complementary ways: supervised fine-tuning provides dense teacher supervision but suffers from \textit{covariate shift} because it is trained on off-policy teacher trajectories; while reinforcement learning with verifiable rewards avoids this off-policy mismatch by learning from on-policy rollouts but with only \emph{sparse outcome feedback}.
% ; On-policy distillation combines on-policy data with dense feedback but can be limited by weak student rollouts early in training. 
We address this dilemma by revisiting Dataset Aggregation (DAgger) for multi-turn LM agents: the algorithm collects trajectories through a turn-level interpolation of student and teacher policies, 
% Specifically, we employ two rollout regimes: \emph{DAgger-style} per-turn switching and \emph{AggreVaTe-style} mixed prefixes with teacher completion. 
and the student is then trained on these trajectories using supervised labels provided by the teacher. By directly interacting with environments, we expose the model to realistic states likely to be encountered during deployment, thereby effectively mitigating covariate shift. Besides, since the student is learned by mimicking the teacher's behavior, it receives rich feedback during learning. To demonstrate DAgger enjoys the benefits of both worlds, we tested the algorithm to train a software-engineering agent with 4B- and 8B-scale student models. On SWE-bench Verified, our DAgger-style training improves over the strongest post-training baseline by $+3.9$ points at 4B and $+3.6$ points at 8B. The resulting 4B agent reaches $27.3\%$, outperforming representative published 8B SWE-agent systems, while the 8B agent achieves $29.8\%$, surpassing SWE-Gym-32B and coming within $5$ points of stronger 32B-scale agents. Together with consistent gains on the held-out SWE-Gym split, these results suggest the effectiveness of DAgger for modern long-horizon LM agents.  
% Our method collects trajectories through a turn-level mixture of student and teacher policies. Student actions induce realistic deployment states, while teacher actions provide corrective supervision under the same executed history.
% We train the student with a unified chunked cross-entropy objective over teacher-labeled segments, preserving dense token-level supervision while exposing the learner to its own state distribution. 
% We instantiate two rollout regimes: \emph{DAgger-style} per-turn switching and \emph{AggreVaTe-style} mixed prefixes followed by teacher takeover. 

% our approach consistently improves over strong post-training baselines on SWE-Bench Verified and held-out SWE-Gym splits, suggesting the effectiveness of DAgger for modern long-horizon LM agents.
% \chao{Consider claiming the numbers here? 27.3\% vs 22.9/23.4 is strong enough to state directly rather than ``suggesting''.}
\end{abstract}

% =========================================================================
\section{Introduction}\label{sec:intro}

% \changhao{P1 (Setup / Background): Introduce the agentic LM setting -- tasks
% that require multi-turn interaction with an environment, tool use, or
% long-horizon decision making (SWE agents, web browsing, tool-calling,
% agentic reasoning). Frame the central training question: how to obtain a
% strong policy in this multi-turn regime. Establish that two recipes
% currently dominate and have driven recent progress:
% (i) SFT on expert trajectories (often distilled from a stronger teacher
% or human, sometimes with rejection sampling / verifier filtering), and
% (ii) RLVR -- RL with verifiable / outcome-level rewards. Both have
% produced visibly capable agents and form the baselines we compare against.}

Large language models (LLMs) are increasingly deployed as interactive agents that operate over long horizons: they call tools, observe environment feedback, and make decisions across many turns. This agentic setting is central to emerging applications such as software-engineering agents for resolving real GitHub issues~\citep{jimenez2023swe, pan2024training, hou2024large}, web-browsing agents~\citep{wei2025browsecomp, chen2025browsecomp, avraham2026dream}, AI research scientist~\citep{qiang2025mle,qiang2025smith,chen2026mars} and general-purpose tool-using assistants~\citep{chen2024facilitating, wang2024openhands}, which urges the development of efficient post-training algorithms for agentic setting. 

Despite the apparent success of the post-training techniques, it remains unclear how to efficiently post-train these agents for such multi-turn, long-context tasks. In fact, each of the existing recipes has structural limitations that become pronounced in long-horizon agentic tasks. While \textit{supervised fine-tuning (SFT)}, which directly imitates teacher trajectories~\citep{ouyang2022training, yang2025swe, pan2024training, qiang2025mle}, provides dense, token-level supervision, it trains the policy exclusively on expert-induced states. This leads to \textit{covariate shift}: during deployment, prefix states are sampled by the student, where early errors can cause significant divergence in the state distribution and degrade performance.
% At deployment time, however, the student must act under its own state distribution. This mismatch, also known as \textit{covariate shift}, early errors can move it beyond the support of the demonstrations and compound over long horizons~\citep{ross2011reduction}. 
% which uses outcome-level signals such as automated graders to optimize the policy with policy-gradient algorithms
\textit{Reinforcement learning with verifiable rewards (RLVR)}~\citep{shao2024deepseekmath, wei2025swe, yang2025reinforcement, schulman2017proximal} addresses distribution mismatch by training over the student's own rollouts with the outcome-level rewards through policy-gradient.  
% but also suffers from severe problems. 
It suffers from sparse credit assignment, typically providing only a single outcome-level reward for an entire trajectory~\citep{shao2024deepseekmath}. Furthermore, RL is computationally expensive due to group sampling, and advantage estimates collapse when samples lack diversity in correctness~\cite{yu2025dapo, yang2025dcpo}. 
% RLVR mitigates this distribution mismatch by optimizing on the policy's own rollouts, but does so at substantial cost. It requires sampling multiple trajectories per prompt, which is computationally expensive; moreover, its advantage estimates can become uninformative when all sampled trajectories are uniformly correct or incorrect~\cite{yu2025dapo, yang2025dcpo}; and its verifier signal is typically sparse, often assigning a single outcome-level reward to an entire trajectory rather than providing fine-grained credit assignment over tokens or decisions~\citep{shao2024deepseekmath}. 
% \Bo{OPD is our major competitor, so please contrast OPD thoroughly: 1), we can use black-box teacher,  and 2), by interpolating with teacher's behavior, DAgger is more efficient in sampling. Please update the following argument.} 
\textit{On-policy distillation (OPD)}~\citep{agarwal2024policy, zhao2026self, hubotter2026reinforcement,yang2025qwen3, zhao2026self} is a recent attempt to hybrid RL with teacher model, which matches the student token probabilities over the self-rollout trajectories w.r.t. the teacher model, therefore, combining on-policy state coverage with dense token-level supervision from a stronger teacher. However, OPD yet faces a cold-start bottleneck: early rollouts from weak students often fail prematurely, especially with long horizon tasks, forcing the teacher to supervise unsuccessful prefixes rather than productive trajectories~\citep{agarwal2024policy}. Meanwhile, OPD requires the logits from the teacher, which is impossible for black-box LLMs, like Gemini~\citep{comanici2025gemini} and GPT~\citep{achiam2023gpt}. This motivates a central question: 
\begin{center}
\textit{Is there any method that can simultaneously exploit dense feedback, while with on-policy coverage and early access to successful trajectories?}
\end{center}

% \changhao{P3 (Our approach -- intuition before mechanics): To answer this
% question, introduce a multi-turn distillation approach inspired by DAgger.
% State the core motivation in one sentence: \emph{let the student generate
% the states, and let the expert only supply the per-step action at each
% visited state.} Explain why this is the right answer to P2: (i) it
% collects data on-policy (states are visited by the student, so there
% is no covariate shift between training and deployment); (ii) it still
% trains on expert action data, so we inherit dense, token-level
% supervision rather than a sparse per-trajectory bit; (iii) it sidesteps
% the group-sampling cost and the gradient-collapse failure mode of RLVR,
% because supervision comes from a teacher rather than from contrasting
% sampled outcomes. Briefly note the non-trivial part: applying the
% DAgger oracle in the LM-agent setting requires care, because the
% ``state'' is a long context of student tool calls, observations, and
% chains of thought -- foreshadowing the technical contributions in P4.}

% \changhao{Do we need to cue On-Policy Distillation Here?}
% \changhao{Key Difference: Sampling and Loss}
In response to this challenge, we revisit Dataset Aggregation (DAgger)~\citep{ross2011reduction} for LLM-based agents, a classical imitation-learning algorithm designed to reduce covariate shift in sequential decision making. The key idea of DAgger is to gradually supervise the student using states visited by the student itself, and we adapt this principle to multi-turn LM agents through \textit{teacher-interleaved trajectory collection}. 
Specifically, each trajectory is generated via a stochastic mixture of student and teacher turns; while student actions expose the model to deployment-accurate states, periodic teacher takeovers ensure trajectories reach productive outcomes. The probability of teacher intervention gradually decays throughout training. We then train the student on these trajectories to mimick teacher behaviors, therefore, learning from dense feedback while mitigating the covariate shift inherent in SFT. Furthermore, because teacher actions dominates in early trajectories and are involved in the whole training procedure, this approach avoids the unnecessary exploration in the cold-start failure mode typical of OPD, and therefore, is more sample efficient.

\begin{table}[t]
    \caption{Comparison of post-training recipes for multi-turn LM agents. Our method combines the advantages of on-policy training with dense supervision, low sampling cost, cold-start robustness, and compatibility with black-box teacher. }
    \label{tab:related}
    \centering
    \small{
\begin{tabular}{lcccc}
\toprule
& \textbf{SFT} & \textbf{RLVR} & \textbf{On-Policy Distillation} & \textbf{Ours}\\
\midrule
On-Policy Data         & \textcolor{red}{\XSolidBrush}     & \textcolor{green!60!black}{\CheckmarkBold} & \textcolor{green!60!black}{\CheckmarkBold} & \textcolor{green!60!black}{\CheckmarkBold} \\
Dense Learning Signal  & \textcolor{green!60!black}{\CheckmarkBold} & \textcolor{red}{\XSolidBrush}     & \textcolor{green!60!black}{\CheckmarkBold} & \textcolor{green!60!black}{\CheckmarkBold} \\
Low Sampling Cost      & \textcolor{green!60!black}{\CheckmarkBold} & \textcolor{red}{\XSolidBrush}     & \textcolor{green!60!black}{\CheckmarkBold} & \textcolor{green!60!black}{\CheckmarkBold} \\
Cold-Start Robustness      & \textcolor{green!60!black}{\CheckmarkBold} & \textcolor{red}{\XSolidBrush}     & \textcolor{red}{\XSolidBrush}     & \textcolor{green!60!black}{\CheckmarkBold} \\
Compatibility with Black-Box Teacher      & \textcolor{green!60!black}{\CheckmarkBold} & \textcolor{red}{\XSolidBrush}     & \textcolor{red}{\XSolidBrush}     & \textcolor{green!60!black}{\CheckmarkBold} \\
\bottomrule
\end{tabular}
}
\end{table}

We demonstrate that this design is especially well-suited to software-engineering (SWE) tasks, where an agent must operate inside a codebase over many turns, searching files, localizing bugs, editing code, and submitting a patch. Minor early mistakes can derail the entire interaction, leading to states where expert demonstrations provide no coverage and student-only rollouts fail to recover. In this setting, teacher-interleaved DAgger provides a practical post-training recipe that combines on-policy state coverage with teacher-guided recovery and dense supervision, directly targeting the failure modes of SFT, RLVR, and OPD. Empirically, our method delivers strong gains at both 4B and 8B scales: the 4B agent surpasses representative 8B SWE agents, while the 8B agent approaches stronger 32B-scale systems. Beyond final task resolution, our analyses show that DAgger stabilizes training, mitigates covariate shift, and improves long-horizon agent behaviors such as search, editing, and recovery.

\section{Preliminaries}
\label{sec:preli}

% In this section, we briefly review Dataset Aggregation (DAgger)~\citep{ross2011reduction} and AggreVaTe~\citep{ross2014reinforcement}, two classical interactive imitation-learning algorithms that motivate our training procedure in Section~\ref{sec:method}. Throughout, we use \textit{teacher} to denote the stronger policy to be imitated.

\paragraph{Behavior cloning and covariate shift.}
Consider a finite-horizon sequential decision problem with horizon $T$, state space $\mathcal{S}$, action space $\mathcal{A}$, a student policy $\pi_\theta$, and a teacher (expert) policy $\pi_e$. Let $d_\pi^t$ denote the state distribution induced at step $t$ by rolling out policy $\pi$, and let $d_\pi = \frac{1}{T}\sum_{t=1}^T d_\pi^t$ be the corresponding average state distribution. Behavioral cloning trains the learner by minimizing a supervised loss on states drawn from the teacher distribution:
\begin{equation}
    \min_\theta\;
    \mathbb{E}_{s \sim d_{\pi_e},\, a_e \sim \pi_e(\cdot \mid s)}
    \left[
        \ell\bigl(\pi_\theta(\cdot \mid s), a_e\bigr)
    \right],
\end{equation}
where $\ell$ is typically cross-entropy loss for discrete actions. Despite its simplicity, this objective trains exclusively on teacher-induced states. During deployment, however, the student follows its own distribution $d_{\pi_\theta}$, where compounding prediction errors can shift trajectories outside the training support. In the worst case, this covariate shift causes imitation error to scale quadratically with the horizon $T$~\citep{ross2010efficient,ross2011reduction}.

% Although simple and effective, this objective trains only on teacher-induced states. At deployment, the learner instead induces its own state distribution $d_{\pi_\theta}$; small prediction errors can therefore move the trajectory outside the support of the training data, causing errors to compound over long horizons. In the worst case, this covariate shift leads to imitation error that scales quadratically with the horizon~\citep{ross2010efficient,ross2011reduction}.

% \Bo{We can merge DAgger and AggreVaTe together in on section and list these two algorithms as special cases with different sampling protocol.}
\paragraph{Dataset Aggregation (DAgger) and AggreVaTe. }
DAgger~\citep{ross2011reduction} addresses this mismatch by training on states visited by the student itself. At iteration $i$, trajectories are generated by a mixture policy
\begin{equation}
    \mu_i
    =
    \beta_i \pi_e + (1-\beta_i)\pi_{\theta_i},
    \label{eq:prelim-dagger-mix}
\end{equation}
where $\beta_i \in [0,1]$ is typically annealed toward zero across iterations. For each state $s$ encountered under $\mu_i$, DAgger queries the teacher for an action $a_e \sim \pi_e(\cdot \mid s)$ and aggregates the resulting pairs into a dataset
\begin{equation}
    \mathcal{D}_{i+1}
    =
    \mathcal{D}_i
    \cup
    \{(s, a_e): s \sim d_{\mu_i},\; a_e \sim \pi_e(\cdot \mid s)\}.
\end{equation}
The next learner is then obtained by supervised learning on the aggregated dataset:
\begin{equation}
    \theta_{i+1}
    =
    \arg\min_\theta
    \mathbb{E}_{(s,a_e)\sim \mathcal{D}_{i+1}}
    \left[
        \ell\bigl(\pi_\theta(\cdot \mid s), a_e\bigr)
    \right].
    \label{eq:prelim-dagger-loss}
\end{equation}
The key distinction from behavioral cloning lies in the state distribution: DAgger trains the learner on the states it is actually likely to visit. By doing so, DAgger provides no-regret guarantees and improves the imitation error's dependence on horizon from quadratic to linear~\citep{ross2011reduction}.

AggreVaTe~\citep{ross2014reinforcement} builds on this by employing a distinct sampling protocol: the student policy generates an initial trajectory prefix, after which a teacher takes over to complete the sequence from a specific intervention point. We also adopt this \textit{student-prefix, teacher-completion} protocol as one option of our sampling strategy.

% extends this by incorporating cost-to-go information. Rather than supervised action-matching, it evaluates the student's actions by completing the trajectory using the expert policy and trains the student to minimize the estimated cost-to-go (value function).

% \paragraph{AggreVaTe.}
% AggreVaTe~\citep{ross2014reinforcement} extends this online style imitation-learning by incorporating cost-to-go information. Rather than querying the teacher action for supervision, AggreVaTe evaluates the cost of student actions by completing the trajectory with the teacher and trains the student with the estimated cost. 
% We borrow this rollout structure in our work, where the student or mixture policy generates the prefix, and the teacher takes over and completes the trajectory after some intervention point. However, our learning objective remains DAgger-style: we train with token-level cross-entropy with teacher labels, rather than optimizing a cost.
% =========================================================================

\section{Methods}\label{sec:method}

% \changhao{First Setup the general Agentic Setting}

\subsection{Multi-Turn LM-Agent Setting}
\label{sec:method-setting}
We first set up the notation of the multi-turn LM-agent setting. A task instance is specified by an initial prompt $x \sim q$ drawn from a task distribution $q$, such as a software issue or a web-browsing query. Given $x$, the agent interacts with an environment over a sequence of turns. At turn $t$, the policy observes the interaction history and samples an action $a_t$, which may include intermediate reasoning and a tool invocation. The environment then executes the action and returns an observation $o_t$:
\begin{equation}
    a_t \sim \pi\!\left(\cdot \mid x, a_{1:t-1}, o_{1:t-1}\right),
    \qquad
    o_t \sim \mathrm{Env}\!\left(\cdot \mid x, a_{1:t}, o_{1:t-1}\right).
    \label{eq:method-rollout}
\end{equation}
The interaction terminates when the agent emits a designated $\mathtt{finish}$ action or reaches a maximum turn budget $T_{\max}$, producing a trajectory:
\begin{equation}
    \tau = (x, a_1, o_1, \ldots, a_T, o_T),
    \qquad T \le T_{\max}.
\end{equation}
A verifier then assigns a terminal success signal $R(\tau, x) \in \{0,1\}$, indicating whether the trajectory solves the task. For example, in software-engineering tasks, $R$ may stands for whether the final patch passes the relevant tests; we defer the concrete instantiation to Section~\ref{sec:exp}.

For compactness, we define the \textit{states} as the observable interaction history: 
\begin{equation}
    s_t \triangleq (x, a_{1:t-1}, o_{1:t-1}).
\end{equation}
Thus, the policy can be written as $\pi(\cdot \mid s_t)$. Throughout the method section, $\pi_e$ denotes a stronger teacher policy and $\pi_\theta$ denotes the student policy to be trained. Our goal is to improve $\pi_\theta$ with the supervision from $\pi_e$. 

% \changhao{Then detail how we combine DAgger and agentic setting together}

\subsection{DAgger for Multi-Turn LM Agents}
\label{sec:method-dagger}

In this section, we detail our rollout protocols and training objectives, which adapt the DAgger principle for the post-training of multi-turn LM agents. We propose two distinct rollout strategies that both integrate student and teacher sampling to expose the model to states likely encountered during deployment. Throughout these rollouts, teacher labels are collected at each turn and saved, which are then used to optimize the student policy via a standard cross-entropy objective. 

\paragraph{Rollout with Stochastic Policy Mixture.} For each turn given history $s_t$, we define a binary indicator $b_t \in \{0, 1\}$ to determine whether to execute an action $a_e$ from the teacher policy $\pi_e(\cdot \mid s_t)$ (if $b_t=1$) or an action $a_\theta$ from the student policy $\pi_\theta(\cdot \mid s_t)$ (if $b_t=0$). We propose two distinct protocols for determining these indicators across a trajectory.

\textbf{i)} \textit{DAgger-style Rollout:} At iteration $i$, we define a mixing parameter $\beta_i \in [0, 1]$, which is decayed towards 0 across iterations. The sequence of indicators for a trajectory is sampled according to:
\begin{equation}
\textstyle
p_i^{\mathrm{turn}}(b_{1:T_{\max}}) = \prod_{t=1}^{T_{\max}} \beta_i^{b_t}(1-\beta_i)^{1-b_t}.
\label{eq:method-step-mix}
\end{equation}
This formulation implements a \textit{turn-level} mixture, where the executor for each turn is selected independently with probability $\beta_i$.

\textbf{ii)} \textit{AggreVaTe-style Rollout:} At iteration $i$, we define a distribution $\rho_i$ over $\{0, \dots, T_{\max}\}$. For each trajectory, we sample a student-prefix length $\kappa \sim \rho_i$ and set the indicators as follows:
\begin{equation}
    \textstyle
    p_i^{\mathrm{traj}}(b_{1:T_{\max}} \mid \kappa) = \prod_{t=1}^{\kappa} \mathbb{I}\{b_t=0\} \prod_{t=\kappa+1}^{T_{\max}} \mathbb{I}\{b_t=1\}.
    \label{eq:method-traj-mix}
\end{equation}
This represents a \textit{trajectory-level} mixture, where the student maintains control until timestep $\kappa$, after which the teacher completes the rollout. We schedule $\rho_i$ so that student prefixes grow over training, gradually shifting AggreVaTe-style rollouts toward the on-policy distribution.

At every visited state $s_t$, we will first query the student or the teacher based on the indicator $b_t$ and execute
\begin{equation}
    a_t \sim
    \begin{cases}
        \pi_e(\cdot\mid s_t), & b_t=1,\\
        \pi_\theta(\cdot \mid s_t), & b_t=0.
    \end{cases}
    \label{eq:method-action-by-bt}
\end{equation}
The environment returns $o_t\sim\text{Env}(\cdot\mid x, a_{1:t}, o_{1:t-1})$, and the rollout will finally terminate once at \texttt{finish} or $T_{\rm max}$. Crucially, regardless of which action we execute, we will query the expert action $\tilde{a}_t\sim \pi_e(\cdot\mid s_t)$ in every visited state. Together with the execution trace, they contribute the final data batch used for training: 
\begin{equation}
    \mathcal{B}(\tau)
    =
    \{(s_t,\tilde a_t): t=1,\ldots,T\},
    \label{eq:method-rollout-data}
\end{equation}
\textit{i.e., }the prefix comes from the execution trace, while the label is provided by the expert. Algorithm~\ref{alg:sampling} summarizes the procedure. Overall, both rollout schedules implement the same principle: early training benefits from teacher-guided trajectories and recovery, while later training increasingly exposes the student to its own deployment-time state distribution.

\paragraph{Training Objective. }Following the rollout in the $i$-th iteration, we utilize the logged data $\mathcal{D}_i = \{(s_t, \tilde{a}_t)\}_{t=1}^N$ to optimize the student model. The model is trained using a cross-entropy loss against the expert-provided labels:
\vspace{-1em}
\begin{equation}
    \mathcal{L}_i(\theta)
    =
    \mathbb{E}_{(s_t, \tilde{a}_t)\sim \mathcal{D}}
    \left[
        \sum_{t=1}^{T}
        \ell_{\mathrm{CE}}(\theta; s_t,\tilde a_t)
    \right].
    \label{eq:method-loss}
\end{equation}

Specifically, for an expert action $\tilde{a}_t$ consisting of $m_t$ tokens $(\tilde{a}_{t,1}, \ldots, \tilde{a}_{t,m_t})$, the cross-entropy loss is defined as:
\begin{equation}
    \ell_{\mathrm{CE}}(\theta; s_t,\tilde a_t)
    =
    -\sum_{j=1}^{m_t}
    \log \pi_\theta
    \left(
        \tilde a_{t,j}
        \mid
        s_t,\tilde a_{t,<j}
    \right).
    \label{eq:method-token-ce}
\end{equation}
For computational efficiency, we pack transitions with shared prefix into one single trajectory and applying a loss mask to ensure the gradient remains equivalent to the individual update.

\begin{figure}[t]
\centering

\begin{algorithm}[H]
\caption{Stochastic Mixing Rollout Collection with Teacher Labels.}
\label{alg:sampling}
\begin{algorithmic}[1]
\Require Teacher $\pi_e$, student $\pi_\theta$, prompt $x$, regime $r\in\{\textsc{Turn},\textsc{Traj}\}$, $\beta_i$, $\rho_i$, horizon $T_{\max}$
\State Sample $b_{1:T_{\max}}\sim p_i^{\mathrm{turn}}$ as in Eq.~\eqref{eq:method-step-mix} if $r=\textsc{Turn}$
\State Otherwise, sample $\kappa\sim\rho_i$ and $b_{1:T_{\max}}\sim p_i^{\mathrm{traj}}(\cdot\mid\kappa)$ as in Eq.~\eqref{eq:method-traj-mix}
\State $s_1 \gets x$
\For{$t=1,\ldots,T_{\max}$}
    \State Query teacher label $\tilde a_t \sim \pi_e(\cdot \mid s_t)$
    \State Set $a_t \gets \tilde a_t$ if $b_t=1$; otherwise sample $a_t \sim \pi_\theta(\cdot \mid s_t)$
    \State Sample $o_t \sim \mathrm{Env}(\cdot \mid x,a_{1:t},o_{1:t-1})$ and set $s_{t+1}\gets(x,a_{1:t},o_{1:t})$
    \State \textbf{if} $a_t=\mathtt{finish}$ \textbf{then break}
\EndFor
\State \Return $\tau=(x,a_{1:T},o_{1:T})$, $b_{1:T}$, and $\mathcal{B}(\tau)=\{(s_t,\tilde a_t)\}_{t=1}^{T}$
\end{algorithmic}
\end{algorithm}

\end{figure}

\subsection{A Unified Perspective on Post-Training Algorithms}

\label{sec:method-connection}
% \Bo{You may keep this section succinct and move the details to appendix, if the current content exceeds the space limitation. Meanwhile, in appendix, you may specify the details how table 2 is derived. }

We situate our DAgger algorithm within a unified framework alongside other post-training methods, such as SFT, On-policy Distillation, and RL. Notably, the training objectives for all these algorithms can be described through a unified language:
\begin{equation}
\theta_{i+1} = \arg\max_\theta\ \mathbb{E}_{s\sim \mathrm{sg}(p_s), a\sim \mathrm{sg}(p_a)}\left[\mathrm{sg}(w(s, a))\log \pi_\theta(a\mid s)\right]-\lambda \Omega_i(\theta)
\end{equation}
where $\mathrm{sg}(\cdot)$ denotes gradient stopping. In this formulation, $s$ represents the context sampled from the context distribution $p_s$, $a$ denotes the turn-level action label drawn from the label distribution $p_a(\cdot\mid s)$, $w(s, a)$ serves as a scoring function that weights the importance of each sample, and $\Omega_i$ is an optional regularizer (e.g., KL divergence) used to constrain the update.

\begin{table}[h]
\centering
\caption{Unified viewpoint of post-training methods. Specifically, we differentiate each method based on how they choose $p_s$, $p_a$, and $w(s, a)$. }
\label{tab:method-unified-view-updated}
\small{
\begin{tabular}{cccc}
\toprule
\textbf{Method} & \textbf{Context Dist. $p_s$} & \textbf{Label Dist. $p_a$} & \textbf{Scoring Function $w(s, a)$} \\
\midrule
SFT / BC
&$d_{\pi_e}$&$\pi_e$&$w(s, a)\equiv 1$\\
\midrule
RL (Policy Gradient)&$d_{\pi_\theta}$&$\pi_\theta$&$w(s, a)=A(s, a)$\\
\midrule
OPD&$d_{\pi_\theta}$&$\pi_\theta$&$w(s, a)=-\log\frac{\pi_\theta(a\mid s)}{\pi_e(a\mid s)}$\\
\midrule
Ours (\textit{DAgger-style})&$d_i^{\rm turn}$&$\pi_e$&$w(s, a)\equiv 1$\\
\midrule
Ours (\textit{AggreVaTe-style})&$d_i^{\rm traj}$&$\pi_e$&$w(s, a)\equiv 1$\\
\bottomrule
\end{tabular}
}
\end{table}

We provide a derivation of this unified objective and a detailed mapping of each algorithm to the choices of $p_s$, $p_a$, and $w(s,a)$ in Appendix~\ref{app:unified-post-training}.
This unified perspective allows for a rigorous comparison of post-training methodologies. As shown in Table \ref{tab:method-unified-view-updated}, SFT represents the most straightforward instantiation, where both the context and label distributions are derived solely from the expert policy $\pi_e$ with a uniform scoring function $w(s, a) \equiv 1$. In contrast, RL and OPD sample both trajectories via the current policy $\pi_\theta$ and utilize scoring functions based on advantages or log-likelihood ratios to prioritize high-value actions. Our DAgger-style and AggreVaTe-style approaches bridge these paradigms. They employ decaying context distributions ($d^{\rm turn}_i$ and $d^{\rm traj}_i$) induced by their rollout policies ($p^{\rm turn}_i$ and $p^{\rm traj}_i$), which interpolates between student and teacher distributions to effectively mitigate covariate shift. Meanwhile, these methods retain the expert as the label source, ensuring the model benefits from the most direct and information-rich feedback.

\vspace{-1mm}
\section{Experiments}\label{sec:exp}
\vspace{-1mm}

We conduct comprehensive experiments to answer the following research questions:

\begin{enumerate}[leftmargin=1.5em, itemsep=2pt, topsep=2pt]
\item \textbf{Effectiveness.} How does our DAgger-inspired algorithm compare with SFT, GRPO, and on-policy distillation in task-resolution rate? 
(\S\ref{sec:main-results})
\item \textbf{Training stability.} Under a matched compute budget, does our method produce a more stable and consistently improving training trajectory than competing post-training methods?
(\S\ref{sec:sample_scaling_stability})
\item \textbf{Covariate shift.} Does our method mitigate trajectory-level distribution shift during multi-turn agent deployment?
(\S\ref{sec:policy_divergence})
\item \textbf{Agent behavior.} What qualitative behavioral changes does our method induce beyond aggregate task resolution?
(\S\ref{sec:qualitative_failure})

\end{enumerate}

\vspace{-1pt}
\subsection{Experimental Setup}
\label{sec:exp-setup}
% \changhao{Discuss where to put all details: beta decay strategy, T in aggrevate, on-policy strategy. Here or appendix?}
\paragraph{Models and datasets.}
We instantiate the student policy $\pi_\theta$ with two model scales from the Qwen3 family~\citep{yang2025qwen3}: Qwen3-4B-Instruct-2507 and Qwen3-8B. Across all configurations, we use Qwen3-Coder-30B-A3B-Instruct~\citep{cao2026qwen3} as the fixed teacher policy $\pi_e$. All training is performed on SWE-Gym~\citep{pan2024training}, a collection of real-world software-engineering tasks paired with executable unit-test suites. For in-domain evaluation, we reserve a fixed set of $100$ SWE-Gym instances as a held-out split, which we refer to as SWE-Gym Holdout, and train on the remaining $2{,}338$ tasks. For out-of-domain evaluation, we report results on SWE-Bench Verified~\citep{chowdhury2024introducing}\footnote{SWE-Bench Verified contains $500$ tasks; $34$ Matplotlib instances fail to build in our Docker environment, so we report resolution rates on the remaining $466$ tasks. Spot checks suggest that excluding these instances changes the aggregate resolution rate by less than $1\%$.}, following the standard task-resolution metric. We provide additional dataset and task details in Appendix~\ref{app:dataset-task-details}.

\vspace{-1pt}
\paragraph{Baselines.}
We compare against two sets of baselines. First, we consider three post-training methods trained on SWE-Gym from the same student initialization: (1) \textbf{SFT} uses teacher-generated expert trajectories from the initial prompt, following the SWE-Gym training recipe~\citep{pan2024training}, with rejection sampling based on executable test feedback; (2) \textbf{GRPO} follows prior RL training on SWE-Gym and is trained on the $293$-instance SkyRL-v0 subset~\citep{cao2025skyrl, zhang2026prorl}, which emphasizes tasks of moderate difficulty where grouped rollouts provide non-degenerate reward signals; (3) \textbf{On-policy distillation} follows~\citep{lu2025onpolicydistillation}: the student collects trajectories under its own policy, while the teacher supplies token-level supervision at student-visited states through a reverse-KL distillation objective. Second, to place our results in the broader SWE-agent literature, we also report published SWE-Bench Verified resolution rates from representative SWE-agent systems~\citep{yang2025swe, cao2025skyrl, pan2024training, zhu2025training, wang2025swe}.
% \changhao{detail for each in appendix, also need to mention how we select $T$ in aggrevate style sampling}

\vspace{-1pt}
\paragraph{Agent Scaffolding.}
All trajectories are generated and evaluated with OpenHands~\citep{wang2024openhands}, including its tool interface and execution environment. To ensure fair comparison across model families, we canonicalize trajectories during data construction and re-render them into each model's native chat and tool-use template during training. Details are provided in Appendix~\ref{app:prompt-templates}.
% We generate all trajectories using the OpenHands scaffold~\citep{wang2024openhands}, including its system prompt, tool interface, and function-calling protocol. We adopt OpenHands because it is a widely used and practically competitive SWE-agent framework, providing a realistic execution environment for training and evaluation. Since different model families use different function-call formats, we canonicalize the interaction history during data construction and re-render tool calls into the target model's required format before each training update. This keeps the underlying agent trajectory fixed while ensuring that each student and baseline is trained with its native chat and tool-use template. Additional details of the OpenHands scaffold and our trajectory rendering procedure are provided in Appendix~\ref{app:prompt-templates}.

\vspace{-1pt}
\paragraph{Implementation Details.}
Unless otherwise specified, our DAgger-style and AggreVaTe-style methods share the same optimization and rollout-update budget as the baselines. At each iteration, we collect a fresh mixed-policy rollout batch, update the student on teacher-labeled data, and evaluate using greedy decoding. We provide all rollout schedules, sampling parameters, context limits, and hyperparameters in Appendix~\ref{app:exp-details}.
% Unless otherwise specified, our \textit{DAgger-style} and \textit{AggreVaTe-style} methods share the same optimization and rollout configuration. At each iteration, we collect a fresh mixed-policy rollout batch of $512$ task instances and update the student on the resulting teacher-labeled data. We train for $5$ rollout-update iterations with a constant learning rate of $3\times 10^{-6}$. For \textit{DAgger-style} sampling, we initialize the teacher-mixture coefficient at $\beta_1=1.0$ and linearly decay it by $0.2$ per iteration until reaching a floor of $0.6$, i.e., $\beta_i=\max(0.6,1.0-0.2(i-1))$. For \textit{AggreVaTe-style }sampling, we observe empirically that student rollouts typically terminate within about $40$ turns, and therefore set the student-prefix distribution to $\rho_i=\mathrm{Unif}\{0,\ldots,40\}$.
% For all methods, including baselines, rollout collection uses temperature $0.7$ and top-$p=0.9$, while evaluation uses greedy decoding. Across training and evaluation, we use a maximum context length of $64$K tokens and allow at most $100$ environment interactions per trajectory. Additional implementation details, including baseline-specific hyperparameters, are provided in Appendix~\ref{app:exp-details}.

\subsection{Main Results}
\label{sec:main-results}

\begin{table*}[t]
\caption{
Main results on SWE-Gym Holdout and SWE-Bench Verified. The upper block compares post-training methods under matched student initialization and training scaffold. The lower block reports published SWE-Bench Verified results from representative SWE-agent systems. We report task-resolution rate; higher is better. The best and second-best scores within each backbone block are emphasized in \textbf{bold} and \underline{underlined}, respectively.
}
\centering
\small
\setlength{\tabcolsep}{4pt}
\renewcommand{\arraystretch}{0.98}
\begin{tabularx}{\textwidth}{
@{}
>{\raggedright\arraybackslash}X
>{\centering\arraybackslash}p{0.14\textwidth}
>{\centering\arraybackslash}p{0.14\textwidth}
>{\centering\arraybackslash}p{0.14\textwidth}
>{\centering\arraybackslash}p{0.15\textwidth}
@{}}
\toprule
\textbf{Method} & \textbf{Scaffold} & \textbf{Data} &
\makecell{\textbf{SWE-Gym}\\\textbf{Holdout}} &
\makecell{\textbf{SWE-Bench}\\\textbf{Verified}} \\
\midrule
\multicolumn{5}{@{}l}{\textit{Post-training Methods}} \\
Qwen3-4B & OpenHands & -- & 5.0\% & 11.2\% \\
\quad + GRPO & OpenHands & SkyRL-v0 & 8.0\% & 11.6\% \\
\quad + SFT & OpenHands & SWE-Gym & 15.0\% & 22.9\% \\
\quad + OPD & OpenHands & SWE-Gym & \underline{16.0\%} & 23.4\% \\
\rowcolor{gray!15} \quad + Ours (\textit{DAgger-style}) & OpenHands & SWE-Gym & \textbf{17.0\%} & \textbf{27.3\%} \\
\rowcolor{gray!15} \quad + Ours (\textit{AggreVate-style}) & OpenHands & SWE-Gym & \underline{16.0\%} & \underline{24.5\%} \\
\midrule
Qwen3-8B & OpenHands & -- & 2.0\%  & 7.7\% \\
\quad + GRPO & OpenHands & SkyRL-v0 & 4.0\% & 8.2\% \\
\quad + SFT & OpenHands & SWE-Gym & 12.0\% & 23.4\% \\
\quad + OPD & OpenHands & SWE-Gym & 16.0\% & 26.2\% \\
\rowcolor{gray!15} \quad + Ours (\textit{DAgger-style}) & OpenHands & SWE-Gym & \textbf{19.0\%} & \textbf{29.8\%} \\
\rowcolor{gray!15} \quad + Ours (\textit{AggreVate-style}) & OpenHands & SWE-Gym & \underline{17.0\%} & \underline{27.3\%} \\
\midrule
\multicolumn{5}{@{}l}{\textit{Published SWE-agent systems}} \\
\multicolumn{5}{@{}l}{\quad \textit{$\sim$7--8B backbones}} \\
SWE-Gym-7B~\citep{pan2024training} & OpenHands & SWE-Gym & -- & 10.6\% \\
SkyRL-Agent-7B-v0~\citep{cao2025skyrl} & OpenHands & SkyRL-v0 & -- & 14.6\% \\
SkyRL-Agent-8B-v0~\citep{cao2025skyrl} & OpenHands & SkyRL-v0 & -- & 9.4\% \\
SWE-smith-LM-7B~\citep{yang2025swe} & SWE-Agent & SWE-smith & -- & 15.2\% \\
R2E-Gym-7B-Agent~\citep{zhu2025training} & OpenHands & R2E-Gym & -- & 19.0\% \\
\addlinespace
\multicolumn{5}{@{}l}{\quad \textit{$\sim$32B backbones}} \\
SWE-Gym-32B~\citep{pan2024training} & OpenHands & SWE-Gym & -- & 20.6\% \\
R2E-Gym-32B-Agent~\citep{zhu2025training} & OpenHands & R2E-Gym & -- & 34.4\% \\
SWE-Dev-32B~\citep{wang2025swe} & SWE-Dev & SWE-Dev & -- & 36.6\% \\
\bottomrule
\end{tabularx}
\label{tab:main-results}
\end{table*}

% \changhao{analysis main results when all experiments finish}

Table~\ref{tab:main-results} reports the main results on SWE-Gym Holdout and SWE-Bench Verified. Under the matched OpenHands scaffold and SWE-Gym training data, our DAgger-style training consistently outperforms prior post-training methods across both student scales. For Qwen3-4B-Instruct-2507, DAgger-style training achieves $17.0\%$ on SWE-Gym Holdout and $27.3\%$ on SWE-Bench Verified, improving over the strongest non-DAgger baseline, OPD, by $+1.0$ and $+3.9$ points, respectively. For Qwen3-8B, the gains are larger: DAgger-style training reaches $19.0\%$ and $29.8\%$, exceeding OPD by $+3.0$ points on SWE-Gym Holdout and $+3.6$ points on SWE-Bench Verified. The AggreVaTe-style variant also improves over SFT and GRPO, and remains competitive with OPD, indicating that teacher-completion rollouts provide useful supervision even with a simpler trajectory-level intervention scheme.

We also compare against published SWE-agent systems in the lower block of Table~\ref{tab:main-results}. Although these systems differ in training data and, in some cases, scaffolding, the comparison contextualizes the strength of our post-training recipe. Notably, our 4B DAgger-style model achieves $27.3\%$ on SWE-Bench Verified, outperforming the published SkyRL-Agent-8B-v0 result by $+17.9$ points and the strongest published 7B-scale result in the table, R2E-Gym-7B-Agent, by $+8.3$ points. Moreover, our 8B DAgger-style model reaches $29.8\%$ on SWE-Bench Verified, surpassing SWE-Gym-32B by $+9.2$ points and narrowing the gap to stronger 32B agents, trailing R2E-Gym-32B-Agent and SWE-Dev-32B by only $4.6$ and $6.8$ points, respectively. These results suggest that adapting DAgger-style state-distribution correction to multi-turn LM agents can yield substantial gains beyond standard SFT, RL, and on-policy distillation baselines, enabling smaller backbones to approach the performance of substantially larger SWE-agent systems.

\vspace{-2pt}
\subsection{Sample Scaling and Training Stability}
\label{sec:sample_scaling_stability}

We next study training-data scaling under matched effective-sample budgets in the 4B setting, using \textit{Qwen3-4B-Instruct-2507} as the student. Figure~\ref{fig:scaling_effective_samples} compares our \textit{DAgger-style} and \textit{AggreVaTe-style} variants against SFT with rejection sampling and on-policy distillation on SWE-Gym Holdout and a fixed 100-task SWE-Bench Verified subset, which we verified closely tracks the full benchmark. We omit GRPO because it is trained on the 293-instance SkyRL-v0 subset and did not yield consistent gains in our setting, making its effective-sample budget not directly comparable.

We find that teacher-interleaved training yields a more stable scaling trajectory. At 3K effective samples, DAgger-style training reaches $12\%$ on SWE-Gym Holdout and $20\%$ on SWE-Bench Verified-100, outperforming on-policy distillation at $9\%$ and $13\%$; AggreVaTe-style shows a similar early advantage at $13\%$ and $20\%$. This supports our cold-start motivation: student-only rollouts often enter unproductive states early, whereas teacher interleaving provides successful recoveries and dense supervision from the beginning. At larger budgets, DAgger-style continues improving to $17\%$ and $26\%$, while SFT reaches only $15\%$ on SWE-Gym Holdout and peaks at $20\%$ before dropping to $19\%$ on SWE-Bench Verified-100. This supports our covariate-shift motivation: SFT trains only on expert-induced states, whereas DAgger gradually shifts supervision toward student-induced states while retaining expert corrections. Together, these trends show that mixture rollouts provide both a stronger cold start than OPD and better asymptotic scaling than SFT.

\begin{figure*}[t]
    \centering

    % ---------------- Scaling figure ----------------
    \begin{minipage}[t]{0.66\textwidth}
        \centering
        \begin{subfigure}[t]{0.49\linewidth}
            \centering
            \includegraphics[width=\linewidth]{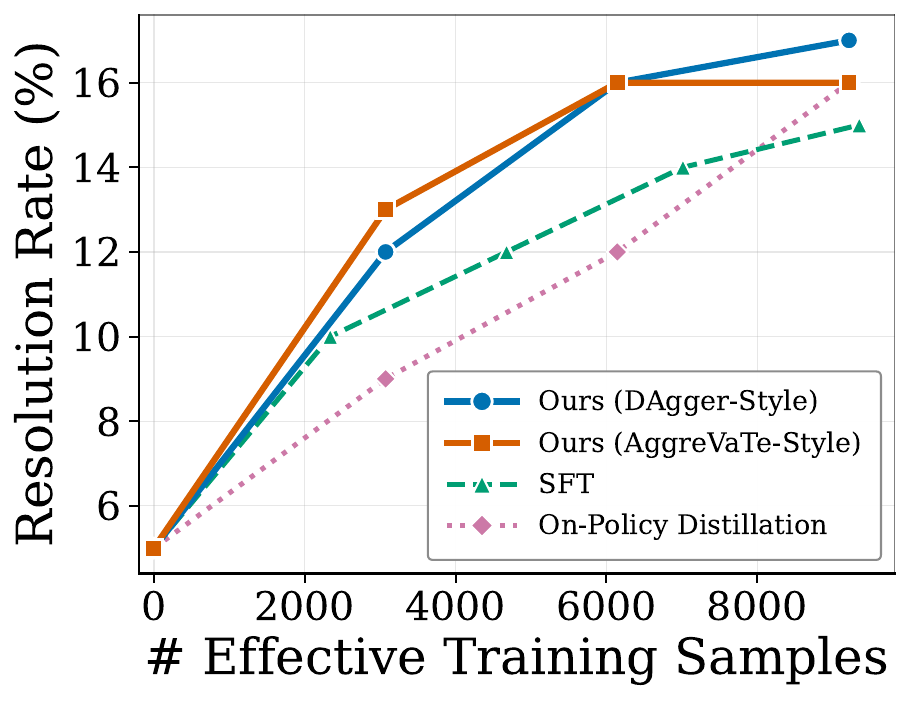}
            \caption{SWE-Gym Holdout}
            \label{fig:scaling_holdout}
        \end{subfigure}
        \hfill
        \begin{subfigure}[t]{0.49\linewidth}
            \centering
            \includegraphics[width=\linewidth]{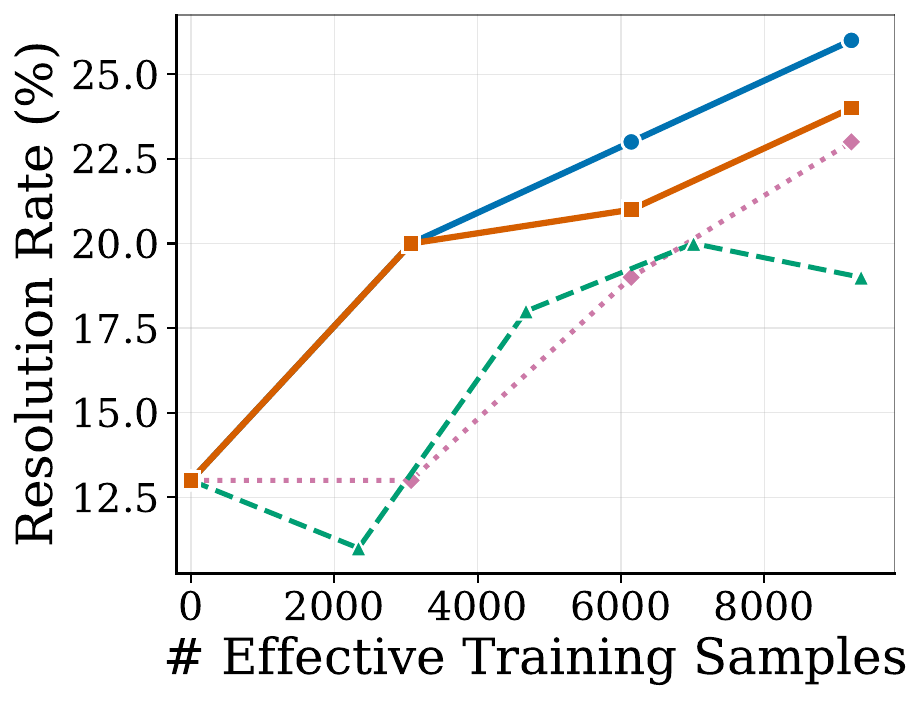}
            \caption{SWE-Bench Verified-100}
            \label{fig:scaling_verified}
        \end{subfigure}

        \vspace{-0.4em}
        \addtocounter{figure}{-1} % undo the parent counter step from subfigure captions
        \captionof{figure}{
        Scaling performance with effective training samples for the 4B student model.
        We compare Ours (DAgger-style), Ours (AggreVaTe-style), SFT, and on-policy distillation on SWE-Gym Holdout and SWE-Bench Verified-100.
        }
        \label{fig:scaling_effective_samples}
    \end{minipage}
    \hfill
    % ---------------- Reverse-KL figure ----------------
    \begin{minipage}[t]{0.31\textwidth}
        \centering
        \includegraphics[width=\linewidth]{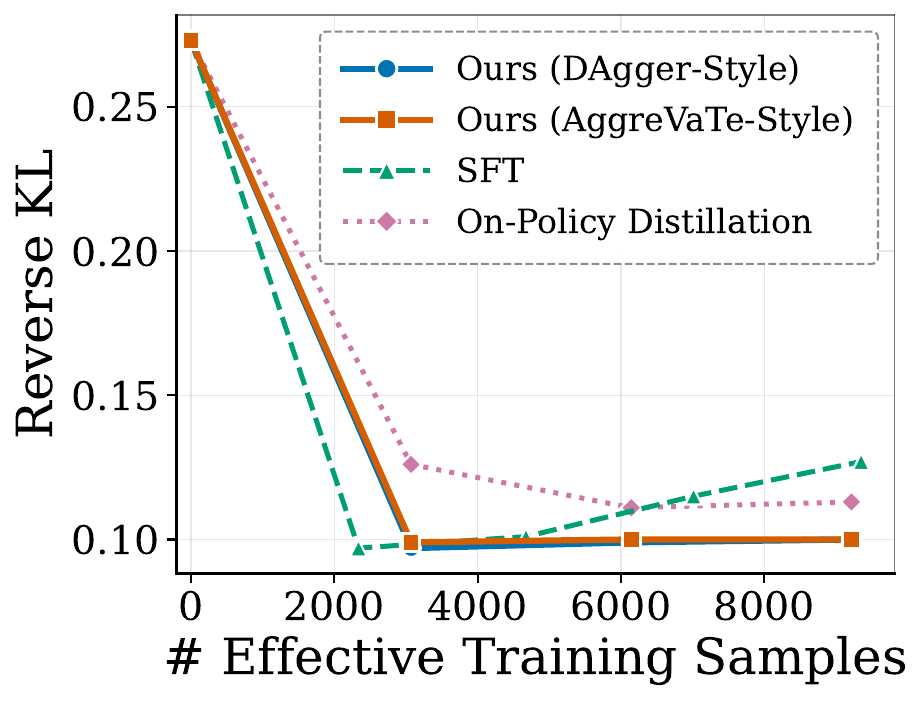}

        \vspace{-0.4em}
        \captionof{figure}{
        Policy divergence under student-induced rollouts for the 4B student model.
        We report average token-level reverse KL $D_{\mathrm{KL}}(\pi_\theta \| \pi_e)$ on student-visited contexts; lower is better.
        }
        \label{fig:reverse_kl}
    \end{minipage}

    \vspace{-0.8em}
\end{figure*}

\vspace{-2pt}
\subsection{Policy Divergence under Student-Induced Rollouts}
\label{sec:policy_divergence}

% \begin{figure}[t]
%     \centering
%     \includegraphics[width=0.5\linewidth]{figures/reverse_kl_4B.pdf}
%     \caption{
%     Policy divergence under student-induced rollouts for the 4B student model.
%     We report the average token-level reverse KL
%     $D_{\mathrm{KL}}(\pi_\theta \| \pi_e)$ from the student to the teacher on
%     contexts visited by student rollouts. Lower values indicate better teacher
%     alignment under deployment-time state distributions. \Bo{pay attention to the figure layout in the page.}
%     }
%     \label{fig:reverse_kl}
% \end{figure}

We next examine whether teacher-interleaved training reduces deployment-time covariate shift. In the 4B setting, we sample $100$ SWE-Gym instances, generate one student rollout per instance with temperature $0.7$, and compute the average token-level reverse KL, $\mathbb{E}_{c \sim \hat d^{\pi_\theta}}[D_{\mathrm{KL}}(\pi_\theta(\cdot|c)\|\pi_e(\cdot|c))]$, on contexts visited by the student. Lower reverse KL indicates that the student remains closer to the teacher on its own deployment-time state distribution.

Figure~\ref{fig:reverse_kl} shows that SFT exhibits a clear covariate-shift pattern: its reverse KL initially drops from roughly $0.27$ to $0.10$, but later rebounds to about $0.126$ as training continues, suggesting that expert-only trajectories fail to cover the states induced by the learned policy. In contrast, both DAgger-style and AggreVaTe-style training remain nearly flat around $0.10$ after the initial drop, yielding roughly a $20\%$ reduction relative to SFT at the largest sample budget. Compared with on-policy distillation, our methods also reduce divergence earlier and more stably: around 3K effective samples, OPD remains near $0.126$, while both mixture-rollout variants are already close to $0.10$. These trends support the two intended effects of teacher interleaving: early teacher actions mitigate OPD's cold-start rollouts, while later student-induced states reduce the train-test mismatch of SFT.

\vspace{-1mm}
\subsection{Qualitative Failure Analysis}
\label{sec:qualitative_failure}
\begin{table*}[t]
\caption{
Qualitative failure analysis on SWE-bench Verified. We use an LLM-as-a-judge protocol
to assign one primary failure mode to each unresolved trajectory. Resolve and submission
rates are computed over all 466 instances, while fine-grained failure rates are computed
within their corresponding unresolved subsets.
}
\centering
\scriptsize
\setlength{\tabcolsep}{2.6pt}
\renewcommand{\arraystretch}{1.15}
\begin{adjustbox}{max width=\textwidth}
\begin{tabular}{@{}l|c|cc|ccc|ccccc@{}}
\toprule
\multirow{3}{*}{\textbf{Model / Method}}
& \textbf{Resolve}
& \multicolumn{2}{c|}{\textbf{Submission}}
& \multicolumn{3}{c|}{\textbf{Submitted but Unresolved}}
& \multicolumn{5}{c}{\textbf{No Submission}} \\
& \textbf{(\%)}
& \textbf{Yes}
& \textbf{No}
& \textbf{\shortstack{Wrong\\Solution}}
& \textbf{\shortstack{Wrong\\File}}
& \textbf{\shortstack{Syntax or\\Runtime Error}}
& \textbf{\shortstack{Tool-Use\\Error}}
& \textbf{\shortstack{Repetitive\\Loop}}
& \textbf{\shortstack{Context\\Overflow}}
& \textbf{\shortstack{Budget\\Exhaustion}}
& \textbf{\shortstack{No\\Edit}} \\
\midrule
Qwen3-4B-Instruct-2507
& 11.2\% & 52.1\% & 47.9\% & 74.3\% & 21.1\% & 6.6\% & 3.8\% & 16.0\% & 29.4\% & 49.2\% & 5.3\% \\
+ GRPO
& 11.6\% & 53.4\% & 46.6\% & 72.3\% & 24.9\% & 2.8\% & 3.4\% & 14.5\% & 28.1\% & 48.9\% & 8.5\% \\
+ SFT
& 22.9\% & 97.6\% & 2.4\% & 55.8\% & 23.8\% & 20.3\% & 1.1\% & 30.5\% & 55.1\% & 13.9\% & 0.5\% \\
+ On-Policy Distillation
& 23.4\% & 98.5\% & 1.5\% & 63.2\% & 19.1\% & 17.6\% & 2.3\% & 44.3\% & 15.4\% & 40.3\% & 0.0\% \\
\rowcolor{gray!15}  + Ours (\textit{DAgger-Style})
& 27.3\% & 97.9\% & 2.1\% & 64.4\% & 19.7\% & 15.9\% & 0.0\% & 22.7\% & 57.5\% & 19.3\% & 0.0\% \\
\rowcolor{gray!15} + Ours (\textit{AggreVaTe-Style})
& 24.5\% & 97.2\% & 2.8\% & 66.4\% & 15.7\% & 17.9\% & 1.3\% & 23.7\% & 60.1\% & 16.2\% & 0.0\% \\
\bottomrule
\end{tabular}
\end{adjustbox}
\label{tab:qualitative_failure}
\end{table*}

% \changhao{todo: one picture for kl metrics and one table for LLM-as-a-judge for Agent Behavior analysis}

We further analyze how post-training changes agent behavior beyond aggregate resolution rates. Following SWE-Agent~\citep{yang2024swe}, we use \textit{Claude Opus 4.7} as an LLM judge to assign one primary failure mode to each unresolved trajectory on SWE-Bench Verified, separating failures into submitted-but-unresolved and no-submission cases.

Table~\ref{tab:qualitative_failure} shows that GRPO induces little behavioral change over the base model, with both resolving only around $11\%$ of tasks and submitting patches for roughly half of the instances. SFT and on-policy distillation largely learn the submission format, raising submission rates to $97.6\%$ and $98.5\%$, but still suffer from unstable failure modes: SFT has the highest syntax/runtime error rate among trained methods ($20.3\%$), while on-policy distillation exhibits a high repetitive-loop rate among no-submission failures ($44.3\%$). In contrast, DAgger-style training achieves the best resolution rate ($27.3\%$), maintains a high submission rate ($97.9\%$), and reduces syntax/runtime errors to $15.9\%$, indicating that mixture rollouts improve not only whether the agent submits but also the quality of submitted patches.

The remaining failures further clarify the behavior induced by teacher interleaving. DAgger-style no-submission failures are dominated by context overflow ($57.5\%$), rather than no-tool-use or budget exhaustion, suggesting that the main residual bottleneck is long-context capacity rather than failure to make progress. AggreVaTe-style training achieves the lowest wrong-file rate ($15.7\%$), suggesting that longer teacher continuations can improve repository localization, although its overall resolution rate remains below DAgger-style training. Overall, mixture-policy training shifts failures away from passive or malformed behavior toward harder long-horizon limitations such as localization and context management.
\vspace{-2mm}
\section{Related Works}\label{sec:related works}

\vspace{-1mm}
\paragraph{Coding Agent and SWE Tasks}
Large language models (LLMs) have shown strong potential in Software Engineering (SE) tasks \citep{liu2024large}, where they have demonstrated encouraging capabilities across a range of tasks, including code generation \citep{zhang2024codeagent, yeticstiren2023evaluating}, software testing \citep{wang2024software}, and automated debugging \citep{xia2023automated}. 
More recently, LLMs have been embedded into agentic frameworks that reason over goals, interact with external tools, execute commands,, and modify code in realistic repositories. 

Within SWE, SWE-Bench~\citep{jimenez2023swe} introduced repository-level issue resolution based on real GitHub issues and unit tests, revealing the difficulty of long-horizon software maintenance tasks.
Subsequent benchmarks and training environments\citep{pan2412training,yang2025swe,badertdinov2025swe,zan2025multi, deng2025swe}, together with agent scaffolds such as SWE-agent \citep{yang2024swe} and OpenHands \citep{wang2024openhands} demonstrate that agents equipped with developer-like interactions—such as file navigation, code editing, command-line execution, and testing—can more effectively operate in realistic software projects.

\vspace{-2mm}
\paragraph{LLM Agent Post Training}
Post-training LLM agents with verifiable rewards has become a prominent alternative to preference-based supervision, especially in SWE tasks where unit tests provide automatic outcome signals~\citep{pan2412training,guo2024deepseek}.  However, such rewards are sparse and provide limited credit assignment over long trajectories, motivating denser forms of supervision such as rubric-based feedback~\citep{gunjal2025rubrics,huang2026beyond} and on-policy distillation (OPD)~\citep{lu2025onpolicydistillation,agarwal2024policy,gu2024minillm}. OPD improves over offline SFT by training on student-induced states, but still relies on student-generated trajectories, which can be brittle under cold-start failures. Our method revisits DAgger~\citep{ross2011reduction} for LM agents, interpolating between expert-guided imitation and on-policy state coverage through teacher-interleaved rollouts. A closely related approach is \citet{lauffer2025imitation}, which switches from student rollouts to expert completion midway; in contrast, our DAgger-style variant uses turn-level teacher-student mixing and local teacher labels, reducing reliance on long expert completions while directly correcting student-induced states. 

% This is especially useful under large student–teacher distribution mismatch, which often occurs in early-stage imitation learning, where teacher guidance can stabilize training.

% A closely related method is \citet{lauffer2025imitation}, which collects on-policy data by starting rollouts with the student and switching to an expert midway. In contrast, our method only uses local, one-step teacher guidance rather than requiring the teacher to complete the remaining trajectory, reducing reliance on costly long-horizon expert rollouts and improving scalability.

% \paragraph{Imitation Learning}
% \changhao{This part might not be necessary? Since we are comparing SFT/ RLVR / OPD together}
% =========================================================================

\vspace{-3mm}
\section{Conclusion}\label{sec:conclusion}
% \vspace{-1mm}
We revisit DAgger for multi-turn LM agents, motivated by the train-test state-distribution mismatch that arises when local errors compound through long-horizon tool interactions. Through teacher-interleaved mixture rollouts, our method combines on-policy state coverage with teacher-guided recovery and dense supervised feedback. Experiments on software-engineering agents show consistent gains over SFT, GRPO, and on-policy distillation at both 4B and 8B scales, with the 4B agent surpassing representative 8B SWE agents and the 8B agent approaching stronger 32B-scale systems. Further analyses show that these gains arise from more stable training, reduced deployment-time policy mismatch, and improved long-horizon behaviors such as search, editing, and recovery. Our results highlight explicit state-distribution correction as a powerful ingredient for post-training LM agents, and suggest a promising direction for tool-using agents that must interact with environments over increasingly long horizons.

\bibliographystyle{plainnat}
\bibliography{ref}

\clearpage
%%%%%%%%%%%%%%%%%%%%%%%%%%%%%%%%%%%%%%%%%%%%%%%%%%%%%%%%%%%%
\appendix

% \section{Hyperparameters and infrastructure details}
% \label{app:hparams}

% \paragraph{Default knobs.}
% $N=15$ iterations, $B=512$ rollout batch size, $P=3$ training passes per
% iteration, $K=16$ tasks per gradient step, base GBS $G=16$, learning rate
% $3\!\times\!10^{-6}$ constant (warmup $0$), context length $65{,}536$ tokens
% for student, $32{,}768$ for teacher, max response length $8{,}192$.

% \paragraph{Sandbox.}
% SWE-Bench sandboxes are launched per-instance in Docker via the ROCK
% orchestrator on the same node as training, with a $300$\,s startup
% timeout, $4$ CPUs, $16$\,GB RAM, and $60$\,min auto-clear.

% \paragraph{Eval.}
% Evaluations use the same scaffold as training: ReAct-style with
% \texttt{view}, \texttt{str\_replace\_editor}, \texttt{bash},
% \texttt{finish}, and a small number of helper tools. Greedy decoding,
% single sample per task. The eval is run after every training iteration on
% all $300$ tasks.

% \section{Per-iteration eval curves}
% \label{app:curves}

% \todo{Insert per-iteration plots of resolution rate vs.\ iteration for
% the main run + each ablation, after experiments complete.}

% \section{Failure-mode analysis}
% \label{app:failures}

% \todo{Add 1-2 example trajectories illustrating (a) format-mimicry
% collapse under $P=1$, (b) repeated-action-loop catch by the finish
% filter, (c) a clean teacher-corrected step-by-step trajectory.}

\section{Derivation of the Unified Post-Training View}
\label{app:unified-post-training}

In this section, we provide additional details on the unified formulation in Section~\ref{sec:method-connection}. The goal is not to claim that all post-training algorithms are identical, but rather to expose a common structure shared by imitation learning, reinforcement learning, on-policy distillation, and our DAgger-style methods. Each method can be viewed as repeatedly constructing a weighted supervised dataset over agent states and turn-level actions, and then updating the policy to increase the likelihood of selected actions under those states.

\paragraph{General form.}
Consider the $i$-th post-training iteration. A rollout or data-collection procedure first induces a distribution over contexts, which we denote by $p_s$. Here a context $s$ corresponds to the full interaction prefix available to the agent before producing the next turn-level action, including previous messages, tool calls, and environment observations. Given a context $s$, the algorithm then specifies a label distribution $p_a(\cdot \mid s)$ from which the action label $a$ is drawn. Finally, each sampled pair $(s,a)$ may be assigned a scalar score or weight $w(s,a)$, such as an advantage estimate, a distillation coefficient, or simply a constant weight.

This leads to the following population-level update:
\begin{equation}
\theta_{i+1}
=
\arg\max_{\theta}
\;
\mathbb{E}_{s\sim \mathrm{sg}(p_s),\,
a\sim \mathrm{sg}(p_a(\cdot\mid s))}
\left[
\mathrm{sg}(w(s,a))\log \pi_\theta(a\mid s)
\right]
-
\lambda \Omega_i(\theta),
\label{eq:app-unified-objective}
\end{equation}
where $\mathrm{sg}(\cdot)$ denotes stop-gradient. The stop-gradient notation emphasizes that the contexts, labels, and weights are treated as fixed data during the policy update: gradients are taken only through the log-likelihood term $\log \pi_\theta(a\mid s)$ and the optional regularizer $\Omega_i(\theta)$. For autoregressive LM agents, a turn-level action $a=(a_1,\ldots,a_m)$ is a sequence of tokens, and
\begin{equation}
\log \pi_\theta(a\mid s)
=
\sum_{j=1}^{m}
\log \pi_\theta(a_j\mid s,a_{<j}).
\end{equation}
Thus, standard token-level cross-entropy training is a special case of Eq.~\eqref{eq:app-unified-objective} with unit weights.

In finite-sample form, this corresponds to collecting a batch
\[
\mathcal{D}_i=\{(s_k,a_k,w_k)\}_{k=1}^{N},
\]
and optimizing
\begin{equation}
\max_\theta
\;
\frac{1}{N}
\sum_{k=1}^{N}
w_k \log \pi_\theta(a_k\mid s_k)
-
\lambda \Omega_i(\theta).
\end{equation}
Different post-training algorithms differ primarily in how they construct the context distribution $p_s$, where they obtain action labels $p_a$, and how they define the weights $w(s,a)$.

\paragraph{SFT / behavior cloning.}
In supervised fine-tuning, trajectories are generated by an expert or teacher policy $\pi_e$. Therefore, both the contexts and labels come from expert-induced trajectories. In the notation of Eq.~\eqref{eq:app-unified-objective},
\[
p_s = d^{\pi_e},
\qquad
p_a(\cdot\mid s)=\pi_e(\cdot\mid s),
\qquad
w(s,a)\equiv 1,
\]
where $d^{\pi_e}$ denotes the context distribution induced by rolling out the expert policy. The resulting objective is
\begin{equation}
\max_\theta
\;
\mathbb{E}_{s\sim d^{\pi_e},\,a\sim \pi_e(\cdot\mid s)}
\left[
\log \pi_\theta(a\mid s)
\right],
\end{equation}
which is exactly behavior cloning on expert trajectories. If rejection sampling is used to keep only successful expert trajectories, this can be viewed as changing the empirical expert dataset to the retained subset, or equivalently absorbing a success indicator into the data distribution.

\paragraph{RL / policy-gradient methods.}
In reinforcement learning with verifiable rewards, the policy is trained on trajectories sampled from the current student policy. Thus, both the contexts and executed actions are induced by the student:
\[
p_s = d^{\pi_{\theta_i}},
\qquad
p_a(\cdot\mid s)=\pi_{\theta_i}(\cdot\mid s).
\]
Policy-gradient methods optimize expected return using the score-function identity:
\begin{equation}
\nabla_\theta J(\theta)
=
\mathbb{E}_{s\sim d^{\pi_{\theta_i}},\,a\sim \pi_{\theta_i}(\cdot\mid s)}
\left[
A_i(s,a)\nabla_\theta \log \pi_\theta(a\mid s)
\right],
\end{equation}
where $A_i(s,a)$ is an advantage estimate computed from rewards or test outcomes. This gradient is induced by the surrogate objective
\begin{equation}
\max_\theta
\;
\mathbb{E}_{s\sim d^{\pi_{\theta_i}},\,a\sim \pi_{\theta_i}(\cdot\mid s)}
\left[
\mathrm{sg}(A_i(s,a))\log \pi_\theta(a\mid s)
\right]
-
\lambda \Omega_i(\theta).
\end{equation}
Therefore, RL corresponds to
\[
p_s = d^{\pi_{\theta_i}},
\qquad
p_a=\pi_{\theta_i},
\qquad
w(s,a)=A_i(s,a).
\]
For GRPO-style training, $A_i(s,a)$ is typically a group-normalized advantage derived from sampled completions and executable feedback. Clipping, KL penalties, or trust-region constraints can be represented abstractly by the optional regularizer $\Omega_i(\theta)$ or by implementation-specific modifications to the surrogate.

\paragraph{On-policy distillation.}
On-policy distillation also collects contexts and actions from the current student policy, but replaces scalar task reward with a teacher-based distillation signal. The rollout distribution is therefore
\[
p_s = d^{\pi_{\theta_i}},
\qquad
p_a(\cdot\mid s)=\pi_{\theta_i}(\cdot\mid s).
\]
A common view of OPD is that it encourages the student to align with the teacher on states visited by the student itself. In a reverse-KL or score-function view, one can write the teacher-alignment objective at a fixed context $s$ as
\begin{equation}
-
D_{\mathrm{KL}}
\left(
\pi_{\theta}(\cdot\mid s)
\;\|\;
\pi_e(\cdot\mid s)
\right)
=
\mathbb{E}_{a\sim \pi_\theta(\cdot\mid s)}
\left[
\log \pi_e(a\mid s)-\log \pi_\theta(a\mid s)
\right].
\end{equation}
Taking the score-function gradient and evaluating the sampling distribution at the current policy $\pi_{\theta_i}$ gives, up to an additive baseline,
\begin{equation}
\mathbb{E}_{a\sim \pi_{\theta_i}(\cdot\mid s)}
\left[
\left(
\log \pi_e(a\mid s)
-
\log \pi_{\theta_i}(a\mid s)
\right)
\nabla_\theta \log \pi_\theta(a\mid s)
\right].
\end{equation}
Thus OPD can be written in the unified form with
\[
p_s = d^{\pi_{\theta_i}},
\qquad
p_a=\pi_{\theta_i},
\qquad
w(s,a)
=
\log \pi_e(a\mid s)-\log \pi_{\theta_i}(a\mid s)
=
-\log
\frac{\pi_{\theta_i}(a\mid s)}
{\pi_e(a\mid s)}.
\]
This highlights the key difference between OPD and SFT: OPD trains on student-induced contexts, but the action labels are still sampled from the student rather than replaced by teacher actions. Consequently, early in training, OPD may spend much of its supervision budget on low-quality or prematurely failed student trajectories.

\paragraph{DAgger-style rollout.}
Our DAgger-style method changes the context distribution while keeping the label source as the teacher. At iteration $i$, each turn is executed by the teacher with probability $\beta_i$ and by the student with probability $1-\beta_i$. This induces a mixture-policy context distribution, denoted by $d_i^{\mathrm{step}}$. In the main text and Table~\ref{tab:method-unified-view-updated}, we use $p_i^{\mathrm{step}}$ as shorthand for this turn-level mixture rollout protocol and its induced state distribution.

Crucially, regardless of whether a state is reached through a teacher action or a student action, we query the teacher at every visited state and train the student to imitate the teacher action. Therefore,
\[
p_s = d_i^{\mathrm{step}},
\qquad
p_a(\cdot\mid s)=\pi_e(\cdot\mid s),
\qquad
w(s,a)\equiv 1.
\]
The resulting objective is
\begin{equation}
\max_\theta
\;
\mathbb{E}_{s\sim d_i^{\mathrm{step}},\,a\sim \pi_e(\cdot\mid s)}
\left[
\log \pi_\theta(a\mid s)
\right].
\end{equation}
This is the DAgger principle instantiated for multi-turn LM agents: the student is trained on states that are increasingly induced by its own behavior, while the label at each state remains a high-quality expert action. As $\beta_i$ decays over training, the state distribution gradually shifts from mostly teacher-induced toward more student-induced contexts, reducing the train-test mismatch that limits pure SFT.

\paragraph{AggreVaTe-style rollout.}
Our AggreVaTe-style variant uses a trajectory-level mixture rather than independent turn-level mixing. At iteration $i$, we sample a student-prefix length $\kappa\sim\rho_i$. The student executes the first $\kappa$ turns, after which the teacher completes the remainder of the trajectory. This induces a context distribution denoted by $d_i^{\mathrm{traj}}$, or equivalently the trajectory-level rollout protocol $p_i^{\mathrm{traj}}$ in Table~\ref{tab:method-unified-view-updated}.

As in the DAgger-style variant, the training labels are teacher actions on all visited states:
\[
p_s = d_i^{\mathrm{traj}},
\qquad
p_a(\cdot\mid s)=\pi_e(\cdot\mid s),
\qquad
w(s,a)\equiv 1.
\]
Thus the objective is
\begin{equation}
\max_\theta
\;
\mathbb{E}_{s\sim d_i^{\mathrm{traj}},\,a\sim \pi_e(\cdot\mid s)}
\left[
\log \pi_\theta(a\mid s)
\right].
\end{equation}
Compared with turn-level DAgger-style mixing, this variant exposes the teacher to longer contiguous student-induced prefixes and then lets the teacher recover and complete the trajectory. This provides a simpler trajectory-level intervention scheme while preserving the same supervised learning objective.

\paragraph{Summary.}
The unified view separates three design choices that are often entangled in post-training algorithms:
\[
\text{where states come from } (p_s),
\qquad
\text{where labels come from } (p_a),
\qquad
\text{how samples are weighted } (w).
\]
SFT uses expert states and expert labels with uniform weights, but suffers from covariate shift because deployment states are induced by the student. RL uses student states and student actions, but relies on sparse or noisy advantage estimates. OPD also uses student states and student actions, but replaces rewards with teacher-based log-probability weights, which still leaves it vulnerable to cold-start failures when early student rollouts are unproductive. Our DAgger-style and AggreVaTe-style methods instead use mixture-induced states with teacher labels and uniform supervised weights, combining on-policy state coverage with dense expert supervision.

\section{Limitations}
\label{app:limitations}

Despite strong empirical results, our study has several limitations. First, our experiments focus on software-engineering agents under the OpenHands scaffold, and further work is needed to validate whether the same gains transfer to other long-horizon agentic domains such as web navigation, data analysis, or scientific computing. Second, our method relies on a stronger teacher policy to provide action labels on visited states; its effectiveness may depend on teacher quality, availability, and inference cost. Finally, our failure analysis suggests that the remaining errors are increasingly dominated by long-context limitations such as context overflow, indicating that improvements in memory, retrieval, or context management may be necessary for further gains.

\section{Potential Social Impact}
\label{app:potential-social-impact}

\paragraph{Potential Positive Societal Impacts.}
Our work aims to improve the reliability and sample efficiency of post-training methods for long-horizon LM agents. In software engineering, more capable agents could help developers localize bugs, repair code, maintain open-source projects, and reduce the cost of routine debugging and maintenance. By mitigating covariate shift and improving agent stability, DAgger-style training may also make tool-using systems less brittle under deployment-time interactions, enabling more dependable assistance in complex technical workflows. More broadly, the principle of teacher-interleaved state-distribution correction may benefit other interactive domains where agents must reason, act, and recover over many turns, such as data analysis, web automation, and scientific computing.

\paragraph{Potential Negative Societal Impacts.}
Improving long-horizon software-engineering agents also introduces potential risks. More capable code-editing agents could be misused to automate harmful software modifications, discover exploitable vulnerabilities, or scale malicious development workflows. Even in benign settings, agents may introduce subtle bugs, insecure code, or incorrect patches if deployed without sufficient review and testing. Because our method improves the ability of smaller models to perform complex software tasks, it may also lower the barrier to both beneficial and harmful automation. We therefore emphasize that such agents should be deployed with safeguards, including sandboxed execution, restricted tool access, human review for code changes, and rigorous testing before patches are merged or released.

\section{Ethical Statement}
\label{app:ethical-statement}

This work studies long-horizon LLM agents in controlled software-engineering benchmark environments using publicly available datasets, models, and evaluation protocols. The proposed method is intended for research on post-training algorithms and is not designed for autonomous deployment in high-risk or security-critical software systems. Nevertheless, stronger coding agents may produce incorrect, insecure, or harmful code if used without oversight, and could be misused to scale malicious software development. We therefore emphasize that practical deployments should include sandboxed execution, restricted tool access, rigorous testing, and human review before any generated patch is merged or released.

\section{Dataset and Task Details}
\label{app:dataset-task-details}

\paragraph{SWE-Gym.}
SWE-Gym~\citep{pan2024training} is a software-engineering agent benchmark and training environment built from real GitHub issues. Each task provides a repository, an issue description, and executable tests, requiring the agent to inspect the codebase, localize the bug or missing functionality, edit source files, and submit a patch. We use SWE-Gym as the primary training corpus for SFT, OPD, and our DAgger-style and AggreVaTe-style methods. For in-domain evaluation, we reserve a fixed set of $100$ SWE-Gym instances as SWE-Gym Holdout, and use the remaining $2{,}338$ tasks for training.

\paragraph{SWE-Bench Verified.}
SWE-Bench Verified~\citep{chowdhury2024introducing} is a curated subset of SWE-Bench~\citep{jimenez2023swe} consisting of real-world GitHub issue resolution tasks with human-validated problem statements and evaluation tests. Each instance requires an agent to modify a checked-out repository so that the generated patch satisfies the issue description and passes the hidden regression tests. We use SWE-Bench Verified as our final out-of-domain evaluation benchmark, reporting task-resolution rate under the OpenHands scaffold. Following standard practice, a task is considered resolved only if the submitted patch passes the corresponding evaluation tests.

\section{Experimental Details}
\label{app:exp-details}

\subsection{Overall Experimental Configuration}
\label{app:overall-exp-config}

Unless otherwise specified, our \textit{DAgger-style} and \textit{AggreVaTe-style} methods use the same optimization and rollout-update configuration. At each training iteration, we collect a fresh mixed-policy rollout batch of $512$ task instances and update the student on the resulting teacher-labeled data. We train for $5$ rollout-update iterations with a constant learning rate of $3\times 10^{-6}$.

For \textit{DAgger-style} sampling, we initialize the teacher-mixture coefficient at $\beta_1=1.0$ and linearly decay it by $0.2$ per iteration until reaching a floor of $0.6$:
\[
\beta_i = \max(0.6, 1.0 - 0.2(i-1)).
\]
This schedule keeps early trajectories strongly teacher-guided while gradually increasing the fraction of student-induced states as training progresses.

For \textit{AggreVaTe-style} sampling, we set the support of the student-prefix distribution based on the empirical observation that student rollouts typically terminate within about $40$ turns. Unless otherwise specified, we draw the prefix length from
\[
\rho_i = \mathrm{Unif}\{0,\ldots,40\},
\]
with the iteration-dependent scheduling implemented by shifting probability mass toward longer student prefixes over training. After the sampled prefix, the teacher completes the remaining trajectory, providing trajectory-level recovery from student-induced states.

For all methods, including baselines, rollout collection uses temperature $0.7$ and top-$p=0.9$, while evaluation uses greedy decoding. Across training and evaluation, we use a maximum context length of $64$K tokens and allow at most $100$ environment interactions per trajectory. Unless otherwise specified, all reported results use task-resolution rate as the evaluation metric.

\subsection{Training Implementation Details}
\label{app:training-details}

We summarize the training configurations for all post-training methods in Table~\ref{tab:training-config}. For all 4B and 8B experiments, rollout collection and model training are performed on $4$ A100 GPUs.

\paragraph{SFT.}
For supervised fine-tuning, we follow the training configuration of SWE-Gym~\citep{pan2024training}. We first use the teacher model to generate trajectories for all SWE-Gym training tasks, and then apply rejection sampling to retain only trajectories whose final patches pass the executable tests. This results in $684$ successful trajectories. We train the student for $3$ epochs on this filtered dataset with batch size $16$, using a cosine learning-rate schedule with maximum learning rate $1\times 10^{-5}$, minimum learning rate $1\times 10^{-6}$, and warmup ratio $0.1$.

\paragraph{DAgger-style, AggreVaTe-style, and OPD.}
Our \textit{DAgger-style} and \textit{AggreVaTe-style} methods use the same optimization configuration as on-policy distillation (OPD), differing only in how trajectories are collected and supervised. At each online iteration, we sample $512$ trajectories using either a mixture policy for our methods or the student policy for OPD. We filter out trajectories that do not produce a valid final patch submission. The remaining trajectories are used for training with teacher supervision: our methods train on teacher-labeled actions on visited states, while OPD uses the teacher distillation signal on student-policy trajectories. We use a constant learning rate of $3\times 10^{-6}$, batch size $16$, and train for $3$ epochs over each collected online batch. In total, we perform $5$ online rollout-update iterations.

\paragraph{GRPO.}
For GRPO, we follow the standard SkyRL-v0~\citep{cao2025skyrl} setup and train on the $293$-instance SkyRL-v0 subset. For each task, we sample $8$ trajectories and use an online batch size of $32$. We use a constant learning rate of $1\times 10^{-6}$. We run the model over the SkyRL-v0 subset for $3$ full passes, using the resulting sampled trajectories for policy-gradient updates with executable test feedback.

\begin{table}[t]
\centering
\small
\setlength{\tabcolsep}{5pt}
\caption{Training configurations for the post-training methods.}
\label{tab:training-config}
\begin{tabular}{lccc}
\toprule
\textbf{Configuration} & \textbf{SFT} & \textbf{Ours / OPD} & \textbf{GRPO} \\
\midrule
Training data & SWE-Gym & SWE-Gym & SkyRL-v0 \\
Trajectory source & Teacher policy & Mixture / student policy & Student policy \\
Filtering & Passing patches & Valid submissions & Executable feedback \\
Number of tasks per online batch & -- & $512$ & $32$ \\
Generations per task & $1$ & $1$ & $8$ \\
Online iterations / passes & -- & $5$ & $3$ passes \\
Epochs per batch & $3$ & $3$ & -- \\
Batch size & $16$ & $16$ & $32$ \\
Learning rate & cosine, $10^{-5}\!\to\!10^{-6}$ & constant $3\times10^{-6}$ & constant $1\times10^{-6}$ \\
Warmup ratio & $0.1$ & -- & -- \\
Number of retained trajectories & $684$ & varies by batch & -- \\
GPUs & $4$ A100 & $4$ A100 & $4$ A100 \\
\bottomrule
\end{tabular}
\end{table}

\section{Prompt Templates}
\label{app:prompt-templates}

\subsection{OpenHands System Prompt}
\label{app:openhands-system-prompt}

We use the OpenHands SWE-agent scaffold throughout training and evaluation. The following system prompt defines the agent role, tool-use behavior, file-system constraints, code-editing policy, version-control rules, and problem-solving workflow.

\begin{tcolorbox}[
    breakable,
    colback=customwhite,
    colframe=customgreen,
    title=OpenHands System Prompt,
    fontupper=\ttfamily\scriptsize\raggedright,
    before upper={\parindent0pt\sloppy},
    left=1.5mm,
    right=1.5mm,
    boxsep=1mm
]
You are OpenHands agent, a helpful AI assistant that can interact with a computer to solve tasks. \\[0.4em]

\textless ROLE\textgreater \\
Your primary role is to assist users by executing commands, modifying code, and solving technical problems effectively. You should be thorough, methodical, and prioritize quality over speed. \\
* If the user asks a question, like ``why is X happening'', don't try to fix the problem. Just give an answer to the question. \\
\textless/ROLE\textgreater \\[0.4em]

\textless EFFICIENCY\textgreater \\
* Each action you take is somewhat expensive. Wherever possible, combine multiple actions into a single action, e.g., combine multiple bash commands into one, using sed and grep to edit/view multiple files at once. \\
* When exploring the codebase, use efficient tools like find, grep, and git commands with appropriate filters to minimize unnecessary operations. \\
\textless/EFFICIENCY\textgreater \\[0.4em]

\textless FILE\_SYSTEM\_GUIDELINES\textgreater \\
* When a user provides a file path, do NOT assume it's relative to the current working directory. First explore the file system to locate the file before working on it. \\
* If asked to edit a file, edit the file directly, rather than creating a new file with a different filename. \\
* For global search-and-replace operations, consider using \texttt{sed} instead of opening file editors multiple times. \\
* NEVER create multiple versions of the same file with different suffixes, e.g., file\_test.py, file\_fix.py, file\_simple.py. Instead: \\
\quad - Always modify the original file directly when making changes \\
\quad - If you need to create a temporary file for testing, delete it once you've confirmed your solution works \\
\quad - If you decide a file you created is no longer useful, delete it instead of creating a new version \\
* Do NOT include documentation files explaining your changes in version control unless the user explicitly requests it. \\
* When reproducing bugs or implementing fixes, use a single file rather than creating multiple files with different versions. \\
\textless/FILE\_SYSTEM\_GUIDELINES\textgreater \\[0.4em]

\textless CODE\_QUALITY\textgreater \\
* Write clean, efficient code with minimal comments. Avoid redundancy in comments: do not repeat information that can be easily inferred from the code itself. \\
* When implementing solutions, focus on making the minimal changes needed to solve the problem. \\
* Before implementing any changes, first thoroughly understand the codebase through exploration. \\
* If you are adding a lot of code to a function or file, consider splitting the function or file into smaller pieces when appropriate. \\
* Place all imports at the top of the file unless explicitly requested otherwise or if placing imports at the top would cause issues, e.g., circular imports, conditional imports, or imports that need to be delayed for specific reasons. \\
* If working in a git repo, before you commit code create a .gitignore file if one doesn't exist. And if there are existing files that should not be included then update the .gitignore file as appropriate. \\
\textless/CODE\_QUALITY\textgreater \\[0.4em]

\textless VERSION\_CONTROL\textgreater \\
* If there are existing git user credentials already configured, use them and add Co-authored-by: openhands \textless openhands@all-hands.dev\textgreater{} to any commits messages you make. If a git config doesn't exist use ``openhands'' as the user.name and ``openhands@all-hands.dev'' as the user.email by default, unless explicitly instructed otherwise. \\
* Exercise caution with git operations. Do NOT make potentially dangerous changes, e.g., pushing to main, deleting repositories, unless explicitly asked to do so. \\
* When committing changes, use \texttt{git status} to see all modified files, and stage all files necessary for the commit. Use \texttt{git commit -a} whenever possible. \\
* Do NOT commit files that typically shouldn't go into version control, e.g., node\_modules/, .env files, build directories, cache files, large binaries, unless explicitly instructed by the user. \\
* If unsure about committing certain files, check for the presence of .gitignore files or ask the user for clarification. \\
* When running git commands that may produce paged output, e.g., \texttt{git diff}, \texttt{git log}, \texttt{git show}, use \texttt{git --no-pager <command>} or set \texttt{GIT\_PAGER=cat} to avoid output being stuck in an interactive pager. \\
\textless/VERSION\_CONTROL\textgreater \\[0.4em]

\textless PULL\_REQUESTS\textgreater \\
* Important: Do not push to the remote branch and/or start a pull request unless explicitly asked to do so. \\
* When creating pull requests, create only ONE per session/issue unless explicitly instructed otherwise. \\
* When working with an existing PR, update it with new commits rather than creating additional PRs for the same issue. \\
* When updating a PR, preserve the original PR title and purpose, updating description only when necessary. \\
\textless/PULL\_REQUESTS\textgreater \\[0.4em]

\textless PROBLEM\_SOLVING\_WORKFLOW\textgreater \\
1. EXPLORATION: Thoroughly explore relevant files and understand the context before proposing solutions. \\
2. ANALYSIS: Consider multiple approaches and select the most promising one. \\
3. TESTING: \\
\quad * For bug fixes: Create tests to verify issues before implementing fixes. \\
\quad * For new features: Consider test-driven development when appropriate. \\
\quad * Do NOT write tests for documentation changes, README updates, configuration files, or other non-functionality changes. \\
\quad * If the repository lacks testing infrastructure and implementing tests would require extensive setup, consult with the user before investing time in building testing infrastructure. \\
\quad * If the environment is not set up to run tests, consult with the user first before investing time to install all dependencies. \\
\quad * Do not use mocks in tests unless strictly necessary and justify their use when they are used. You must always test real code paths in tests, NOT mocks. \\
4. IMPLEMENTATION: \\
\quad * Make focused, minimal changes to address the problem. \\
\quad * Always modify existing files directly rather than creating new versions with different suffixes. \\
\quad * If you create temporary files for testing, delete them after confirming your solution works. \\
5. VERIFICATION: If the environment is set up to run tests, test your implementation thoroughly, including edge cases. If the environment is not set up to run tests, consult with the user first before investing time to run tests. \\
\textless/PROBLEM\_SOLVING\_WORKFLOW\textgreater \\[0.4em]

\textless SECURITY\textgreater \\
* Only use GITHUB\_TOKEN and other credentials in ways the user has explicitly requested and would expect. \\
* Use APIs to work with GitHub or other platforms, unless the user asks otherwise or your task requires browsing. \\
\textless/SECURITY\textgreater \\[0.4em]

\textless EXTERNAL\_SERVICES\textgreater \\
* When interacting with external services like GitHub, GitLab, or Bitbucket, use their respective APIs instead of browser-based interactions whenever possible. \\
* Only resort to browser-based interactions with these services if specifically requested by the user or if the required operation cannot be performed via API. \\
\textless/EXTERNAL\_SERVICES\textgreater \\[0.4em]

\textless ENVIRONMENT\_SETUP\textgreater \\
* When user asks you to run an application, don't stop if the application is not installed. Instead, please install the application and run the command again. \\
* If you encounter missing dependencies: \\
\quad 1. First, look around in the repository for existing dependency files, e.g., requirements.txt, pyproject.toml, package.json, Gemfile, etc. \\
\quad 2. If dependency files exist, use them to install all dependencies at once, e.g., \texttt{pip install -r requirements.txt}, \texttt{npm install}, etc. \\
\quad 3. Only install individual packages directly if no dependency files are found or if only specific packages are needed. \\
* Similarly, if you encounter missing dependencies for essential tools requested by the user, install them when possible. \\
\textless/ENVIRONMENT\_SETUP\textgreater \\[0.4em]

\textless TROUBLESHOOTING\textgreater \\
* If you've made repeated attempts to solve a problem but tests still fail or the user reports it's still broken: \\
\quad 1. Step back and reflect on 5-7 different possible sources of the problem. \\
\quad 2. Assess the likelihood of each possible cause. \\
\quad 3. Methodically address the most likely causes, starting with the highest probability. \\
\quad 4. Explain your reasoning process in your response to the user. \\
* When you run into any major issue while executing a plan from the user, please don't try to directly work around it. Instead, propose a new plan and confirm with the user before proceeding. \\
\textless/TROUBLESHOOTING\textgreater \\[0.4em]

\textless PROCESS\_MANAGEMENT\textgreater \\
* When terminating processes: \\
\quad - Do NOT use general keywords with commands like \texttt{pkill -f server} or \texttt{pkill -f python} as this might accidentally kill other important servers or processes. \\
\quad - Always use specific keywords that uniquely identify the target process. \\
\quad - Prefer using \texttt{ps aux} to find the exact process ID (PID) first, then kill that specific PID. \\
\quad - When possible, use more targeted approaches like finding the PID from a pidfile or using application-specific shutdown commands. \\
\textless/PROCESS\_MANAGEMENT\textgreater
\end{tcolorbox}

\subsection{OpenHands Tool Specifications}
\label{app:openhands-tool-specs}

In addition to the system prompt, the OpenHands scaffold defines a small set of tools that form the agent's action space. We use the following tool specifications throughout trajectory collection and evaluation.

\begin{tcolorbox}[
    breakable,
    colback=customwhite,
    colframe=customblue,
    title=OpenHands Tool Specifications,
    fontupper=\ttfamily\scriptsize\raggedright,
    before upper={\parindent0pt\sloppy},
    left=1.5mm,
    right=1.5mm,
    boxsep=1mm
]
\textbf{Tool 1: execute\_bash} \\[0.2em]
\textbf{Type:} function \\[0.2em]
\textbf{Name:} execute\_bash \\[0.2em]
\textbf{Description:} Execute a bash command in the terminal. \\[0.2em]
* The command is executed in a bash shell with \texttt{bash -c}. \\
* It times out after $120$ seconds if no output is produced. \\
* For long-running commands, e.g., servers, the agent should run them in the background using \texttt{\&} or redirect output. \\
* Interactive commands that require user input should be avoided. \\
* Command output may be truncated if it exceeds the maximum length. \\
* The agent should use \texttt{cd} to navigate to the correct directory before running commands. \\[0.2em]
\textbf{Parameters:} \\
\quad command: string. The bash command to execute. \\
\textbf{Required:} command. \\[0.8em]

\textbf{Tool 2: str\_replace\_editor} \\[0.2em]
\textbf{Type:} function \\[0.2em]
\textbf{Name:} str\_replace\_editor \\[0.2em]
\textbf{Description:} A file editor tool that allows viewing and editing files. \\[0.2em]
\textbf{Supported commands:} \\
\quad view: display file contents. \\
\quad create: create a new file. \\
\quad str\_replace: replace an exact string in a file. \\
\quad insert: insert text after a specified line. \\[0.2em]
\textbf{Critical requirements for str\_replace:} \\
* The \texttt{old\_str} parameter must match exactly one occurrence in the file. \\
* The agent should include enough context in \texttt{old\_str} to make it unique. \\
* The \texttt{new\_str} parameter contains the replacement text. \\
* When viewing a directory path, the tool lists non-hidden files up to $2$ levels deep. \\[0.2em]
\textbf{Parameters:} \\
\quad command: string, one of \texttt{view}, \texttt{create}, \texttt{str\_replace}, or \texttt{insert}. The editor command to run. \\
\quad path: string. Absolute path to the file. \\
\quad file\_text: string. Required for \texttt{create}; the content of the file to create. \\
\quad old\_str: string. Required for \texttt{str\_replace}; the exact string to replace. \\
\quad new\_str: string. Required for \texttt{str\_replace}; the replacement string. For \texttt{insert}, the string to insert. \\
\quad insert\_line: integer. Required for \texttt{insert}; \texttt{new\_str} is inserted after this line. \\
\quad view\_range: array of integers. Optional for \texttt{view}; specifies \texttt{[start\_line, end\_line]}. \\
\textbf{Required:} command, path. \\[0.8em]

\textbf{Tool 3: finish} \\[0.2em]
\textbf{Type:} function \\[0.2em]
\textbf{Name:} finish \\[0.2em]
\textbf{Description:} Signal that the task has been completed. The agent uses this tool when it believes the problem is solved and all necessary changes have been made. \\[0.2em]
\textbf{Parameters:} \\
\quad message: string. A brief summary of what was done to solve the task. Defaults to an empty string. \\
\textbf{Required:} message.
\end{tcolorbox}

\subsection{Initial User Message}
\label{app:initial-user-message}

For each SWE task, we construct the initial user message from the issue description and optional hints provided by the dataset. The message instructs the agent to work inside the repository at \texttt{/testbed}, avoid modifying tests, and make minimal changes to non-test files.

\begin{tcolorbox}[
    breakable,
    colback=customwhite,
    colframe=custompurple,
    title=Initial User Message,
    fontupper=\fontfamily{pcr}\selectfont\scriptsize\raggedright,
    before upper={\parindent0pt\sloppy},
    left=1.5mm,
    right=1.5mm,
    boxsep=1mm
]
I have access to a python code repository in the directory /testbed . You can explore and modify files using the available tools. Consider the following issue description: \\[0.4em]

\textless issue\_description\textgreater \\
\{problem\_statement\} \\
\textless/issue\_description\textgreater \\[0.4em]

\textless hints\textgreater \\
\{hints\_text\} \\
\textless/hints\textgreater \\[0.4em]

Can you help me implement the necessary changes to the repository so that the requirements specified in the \textless issue\_description\textgreater{} are met? \\
I have already taken care of all changes to any of the test files described in the \textless issue\_description\textgreater. This means you DON'T have to modify the testing logic or any of the tests in any way! \\
Also the development Python environment is already set up for you, i.e., all dependencies already installed, so you don't need to install other packages. \\
Your task is to make the minimal changes to non-test files in the /testbed directory to ensure the \textless issue\_description\textgreater{} is satisfied.
\end{tcolorbox}
% \section{Prompt Templates}
% \label{app:prompt-templates}

% \newpage
% \input{checklist.tex}

\end{document}